\documentclass[letterpaper, 10 pt, conference]{ieeeconf}  

\IEEEoverridecommandlockouts                              

\usepackage{amsmath}
\usepackage{graphicx}
\usepackage{booktabs} 
\usepackage{multirow}
\usepackage{tabularx}
\usepackage{xcolor}
\usepackage{cite}
\usepackage[colorlinks]{hyperref}
  \hypersetup{
    citecolor=blue,
    linkcolor=blue,   
    urlcolor=blue}

\title{\LARGE \bf
Scalable Multi-Task Data Generation via Reinforcement Learning\\ for Language-Conditioned Bimanual Dexterous Manipulation}

\author{Zechu Li$^{1,5}$, Yufeng Jin$^{1,2}$, Puze Liu$^{3,4}$, 
Jan Peters$^{1,3,5}$, Georgia Chalvatzaki$^{1,5}$
\thanks{Corresponding author: Zechu Li (zechu.li@tu-darmstadt.de).}
\thanks{This work was supported by the German Research Foundation (DFG) Emmy Noether Programme (CH 2676/1-2) and funded by the European Union. Views and opinions expressed are however those of the author(s) only and do not necessarily reflect those of the European Union or the European Research Council Executive Agency. Neither the European Union nor the granting authority can be held responsible for them.}
\thanks{$^{1}$Department of Computer Science, TU Darmstadt, Germany.}
\thanks{$^{2}$Honda Research Institute Europe GmbH, Offenbach, Germany.}
\thanks{$^{3}$DFKI, Research Department SAIROL, Darmstadt, Germany.}
\thanks{$^{4}$Tongji University, Shanghai Research Institute for Intelligent Autonomous Systems.}
\thanks{$^{5}$Hessian.AI, Darmstadt, Germany}
}

\begin{document}

\maketitle

\begin{abstract}

A key bottleneck in training generalist policies for bimanual dexterous manipulation is the lack of large-scale, high-quality datasets. Synthetic data generation in simulation provides a scalable alternative to human video demonstrations by overcoming challenges such as morphology mismatch, missing physical interactions, and the generation of robot actions. However, existing approaches based on human teleoperation offer limited task diversity, as object-centric trajectory matching often neglects the feasibility of robot execution. Reinforcement learning (RL) enables broader scalability but is often constrained by handcrafted, task-specific rewards. In this work, we propose a systematic RL-based data generation pipeline that integrates generalizable reward design, effective domain randomization, and language-conditioned task annotations. This pipeline synthesizes diverse, high-quality datasets for dexterous bimanual manipulation and enables training of language-conditioned multi-task policies. Our experiments show that the generated data significantly improves generalization across three representative manipulation tasks.

\end{abstract}

\begin{figure*}[t]
    \centering
    \includegraphics[width=\linewidth]{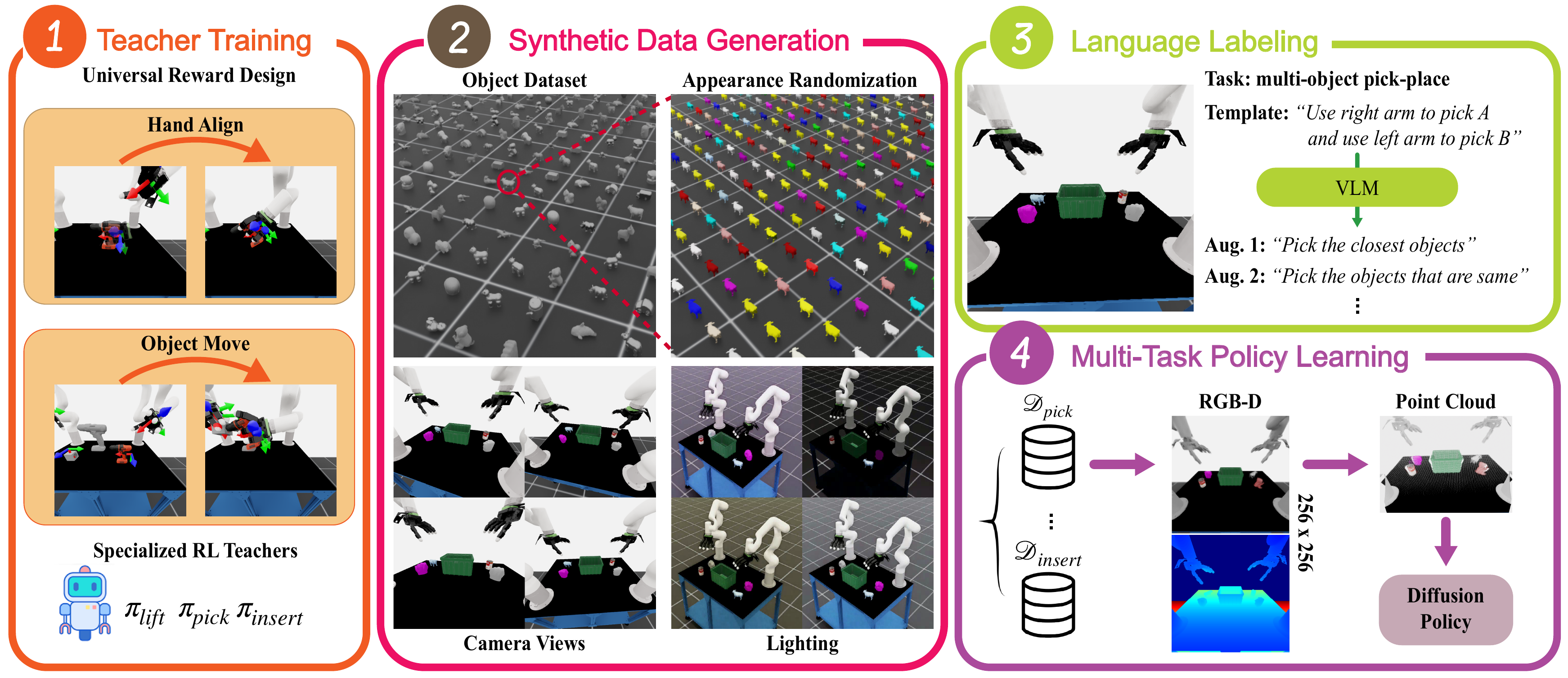}
    \caption{We propose Reinforcement Learning as Data Collector (RLDC), a scalable pipeline for generating synthetic datasets with task-specialized RL teachers and training a language-conditioned multi-task policy for real-world bimanual dexterous manipulation. (1) A universal reward design reduces reward-engineering effort, enabling quick task scaling. (2) Diverse datasets are collected through extensive randomization of objects, physics, and visual conditions. (3) Language labels are automatically generated and further augmented with a VLM for greater diversity. (4) Finally, the collected data are distilled into a visual-based multi-task policy that transfers zero-shot to the real world.}
    \label{fig:overview}
\end{figure*}

\section{INTRODUCTION}
\label{sec:intro}
Bimanual dexterous manipulation is critical for enabling robots to perform complex real-world tasks. 
Recent advances in robotic foundation models, driven by Internet-scale pretraining and grounding with robot actions~\cite{black2024pi_0, team2024octo, kim2024openvla, wen2025tinyvla, reuss2025flower}, have shown impressive capabilities in understanding and executing diverse manipulation skills. However, these models are typically constrained to single-arm setups or bimanual systems with simple parallel grippers, falling short of controlling dexterous hands. This gap largely stems from the absence of large-scale, high-quality, diverse, and multimodal datasets for bimanual dexterous manipulation.

To bridge this gap, multiple data sources have been explored. Human teleoperation offers natural demonstrations~\cite{chi2024diffusionpolicy, fu2024mobile, funk2024actionflow}, but scaling it across diverse tasks and objects is prohibitively expensive and time-consuming. Human videos provide an abundant Internet-scale resource, yet robots face embodiment mismatches, partial observability, and retargeting imperfections that hinder robust policy learning~\cite{qin2023anyteleop, qiu2025humanoid, lum2025crossing, yuan2025hermes}. On the other hand, simulation provides powerful data augmentation capabilities in terms of the visual appearance, objects' diversity, scene complexity, and physics parameters, enabling large-scale dataset collection~\cite{chen2025robotwin, nasiriany2024robocasa, geng2025roboverse, srivastava2022behavior}, but its diversity is still limited and heavily dependent on the quality of expert demonstrations. Prior data augmentation approaches relying on human demonstrations~\cite{jiang2025dexmimicgen, chen2025robotwin, yuan2025hermes} with the assumption of equivariance of object-centric trajectories, often resulting in infeasibility for the robot. For example, when a pot is rotated, the corresponding trajectory is rotated to grasp the handle without considering the complex cluster scene for collision-free execution.

Instead, leveraging Reinforcement Learning (RL) to generate synthetic data offers an attractive alternative: agents can autonomously generate diverse, high-quality trajectories by optimizing rewards and adapting to varied objects, poses, gripper types, and cluttered scenes. Yet, two major challenges hinder its practicality. First, reward engineering requires extensive human effort and must be tailored to each task, limiting scalability on the task level. Second, many RL-based methods rely on explicit state estimation (e.g., object tracking), which makes it challenging for smooth deployment in real-world settings.

To tackle these challenges, we propose Reinforcement Learning as Data Collector (RLDC), a simple yet effective method that leverages reinforcement learning to generate high-quality training data for bimanual dexterous manipulation. RLDC adopts a teacher–student paradigm: we first train per-task state-based teacher policies and then distill them into a single multi-task, diffusion-based student policy. 

While conceptually straightforward, RLDC offers a scalable pipeline for constructing large-scale, multimodal datasets, paving the way for generalist robotic policies in bimanual dexterous manipulation. Our pipeline includes several innovations: (1) in the teacher stage, we \textbf{design a generalizable reward function tailored to bimanual dexterous tasks}, reducing engineering overhead and accelerating policy training; (2) for data collection, we augment the dataset with \textbf{extensive domain randomization together with grounded language task description} over objects' geometry, appearance, physical parameters, poses, lighting, and camera settings, exploiting the fully controlled simulation. The language description of each environment is further expanded using LLM-based augmentation, enriching the language label diversity; (3) finally, we distill the dataset into a \textbf{language-conditioned student policy using diffusion model} to achieve robust bimanual dexterous manipulation. 
To better ground the vision encoder, we introduce an auxiliary loss that anchors visual representations to the underlying state information—an advantage not possible with human teleoperation or video-only data. We evaluate RLDC both in simulation and on real robots and demonstrate that the generalizable reward design achieves success rates comparable to finely tuned task-specific rewards, and that multi-task policies can be learned entirely from synthetic data with strong transfer to the real world.


\section{RELATED WORK}
\label{sec:related}
\subsection{Data Generation for Robot Manipulation}
\label{sec:related:a}
Automated data generation has emerged as a compelling approach for building large-scale robotic datasets~\cite{chen2025robotwin, wang2024cyberdemo, jiang2025dexmimicgen,yuan2025hermes}. Typically, this paradigm starts with a small set of expert demonstrations, augments them through domain randomization, and validates the synthesized trajectories in simulation. For instance, DexMimicGen~\cite{jiang2025dexmimicgen} decomposes tasks into subtasks and generates trajectories for each of them by applying transformations based on the relative pose between the end-effector and the object. RoboTwin 2.0~\cite{chen2025robotwin} leverages LLM agents for policy generation and evaluation, creating a closed-loop architecture for automated data collection. However, the diversity and quality of data produced by these methods remain heavily dependent on the original demonstrations or predefined skills, which significantly constrains their scalability and optimality.

An alternative is to exploit reinforcement learning (RL) policies as teachers in a teacher–student paradigm. Unlike augmented demonstrations, RL teacher policies are optimized under extensive randomization, allowing them to generate more diverse and higher-quality trajectories. Many sim-to-real RL methods adopt this paradigm, but they are typically restricted to specific tasks~\cite{li2025morphologically, lin2024twisting, chen2023visual, qi2023hand, chen2024vegetable}. In contrast, our work investigates its effectiveness in a multi-task setting, while further enriching the data by incorporating language labels to support generalization. Another direction, exemplified by RLDG~\cite{xu2024rldg}, explores directly training RL policies in the real world, thereby circumventing the sim-to-real gap. Although promising for precise manipulation, such real-world RL data generation approaches suffer from low training efficiency and limited scalability compared to simulation-based data generation.

\subsection{Generalizable Reward Design}
\label{sec:related:b}
RL often requires extensive reward engineering for every single task~\cite{eschmann2021reward}. Given the complexity of dexterous manipulation, such manual tuning is tedious and time-consuming, severely limiting scalability across diverse tasks. A natural solution is to design generalizable reward functions that can be applied broadly, thus reducing the need for task-specific engineering. Recent work has explored extracting information from human–object interaction videos, such as human hand poses and object reference trajectories~\cite{yuan2025hermes, dan2025x, chen2025vividex, lum2025crossing}. This enables rewards that generalize across different videos by encouraging robots to imitate hand poses and track object trajectories. However, simply enforcing object trajectory tracking often leads to suboptimal behaviors. For example, when placing an object into a box with randomized object and box poses, naively following the original trajectory may cause collisions with the box's sides. 

A more flexible approach is to define rewards based on contact goals and object goals, allowing the agent to discover feasible trajectories while still achieving the desired final pose~\cite{lin2025sim}. Our method also builds on this idea: instead of relying on contact points, we use reference hand poses as guidance, which ensures functional manipulation while maintaining generalization across tasks.

\subsection{Sim-to-real Transfer for Dexterous Manipulation}
\label{sec:related:c}
A major challenge in sim-to-real transfer arises from visual perception, and three main approaches have been explored for deploying policies in the real world. The first is the state estimation, where object states are inferred through vision pipelines such as segmentation, pose estimation, and tracking, enabling zero-shot execution of state-based policies trained in simulation~\cite{li2025morphologically, lin2024twisting}. While effective in principle, these pipelines are often bottlenecked by high inference latency and suffer from compounded errors due to sensor noise and model inaccuracies, especially in cluttered scenes. The second approach is to use images (RGB and/or depth) as input. Depth images typically exhibit a smaller domain gap than RGB and are therefore widely adopted~\cite{yuan2025hermes, lin2025sim, lum2024dextrah}. To tackle the harder case of RGB input, prior works apply strong augmentation and multi-view learning: e.g., DextrAH-RGB~\cite{singh2024dextrah} leverages stereo RGB with extensive visual randomization, while ManiWhere~\cite{yuan2024learning} employs a multi-view contrastive loss on RGB-D to achieve robust view generalization. Point cloud is an alternative to depth image, which is less constrained by camera-view discrepancies between simulation and reality and allows efficient multi-view fusion, although it requires accurate calibration~\cite{chen2023visual}. Note that, in this work, we adopt point clouds with RGB information as input, combining geometric fidelity with appearance cues to improve robustness in dexterous bimanual manipulation.

\section{METHOD}
\label{sec:method}
We propose Reinforcement Learning as Data Collector (RLDC), a scalable pipeline that leverages task-specialized RL policies to generate diverse, multimodal datasets, which are then distilled into a single multi-task diffusion policy. As illustrated in Fig.~\ref{fig:overview}, RLDC consists of four stages: (1) training task-specific RL teachers using a generalizable reward design; (2) rolling out these teachers in simulation with extensive domain randomization across physical and visual factors; (3) automatically generating corresponding language labels to enrich multimodality; and (4) distilling the collected trajectories into a unified multi-task diffusion policy implemented with a decoupled transformer architecture featuring separate action heads for robust bimanual control.

\subsection{Tasks}

We evaluate our approach on three representative bimanual dexterous manipulation tasks in both simulation and real-world settings: (1) box lifting, (2) multi-object pick-and-place, and (3) drawer insertion. These tasks span different levels of coordination complexity and manipulation patterns, covering a broad range of rigid-body manipulation scenarios. Box lifting represents a single-stage, single-object task that requires precise two-arm coordination and force balance to achieve stable lifting. Multi-object pick-and-place introduces additional challenges in perception and scene understanding, including object detection and stable grasping in cluttered environments and sequential placement while avoiding inter-arm collisions. Drawer insertion further increases complexity by involving articulated object interaction, along with precise timing and bimanual cooperation for successful object insertion. Collectively, these tasks encompass high inter-arm coordination, long-horizon planning, cluttered scene reasoning, and interaction with articulated objects, which provide a comprehensive and challenging testbed for validating the effectiveness and generalization capability of our method.

\begin{figure}[t]
    \centering
    \includegraphics[width=.95\linewidth]{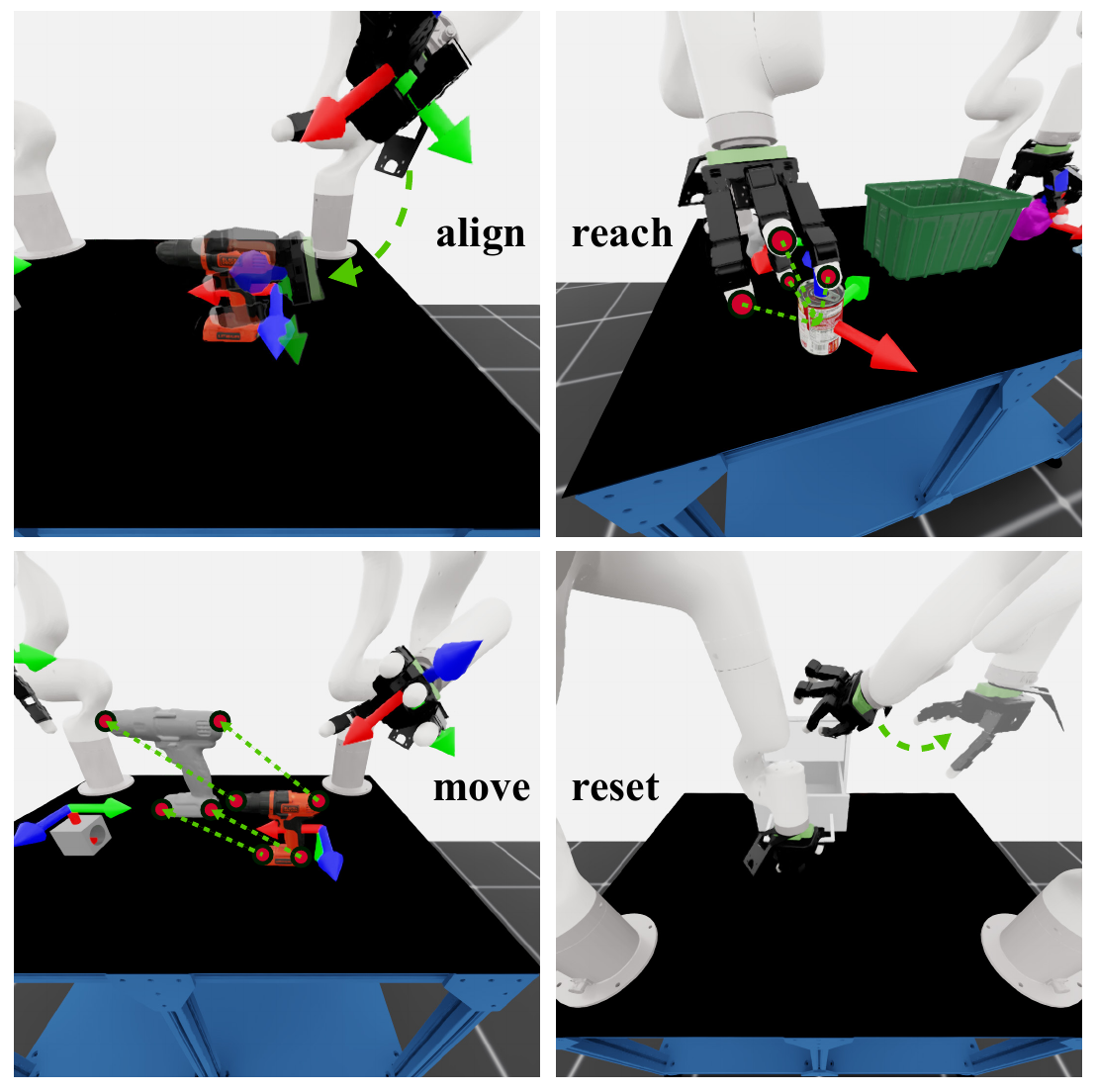}
    \caption{Four reward terms. (Top-Left) A predefined hand pose associated with the drill guides the hand to align accordingly. (Top-Right) The distance between the fingertips, marked by red points, and the object center is minimized to encourage contact. (Bottom-Left) The object is rewarded for moving toward a defined goal pose in the air. (Bottom-Right) After task completion, e.g., placing the object into the drawer, the robot is encouraged to return to a reset configuration.
    }
    \label{fig:reward}
\end{figure}
\subsection{Generalizable Reward Design}
\label{sec:method:a}
RL often relies on carefully designed reward functions that are specific to each task, which hinders its scalability especially for challenging bimanual manipulation with multifingered hands. It is critical to construct a set of generalizable reward terms that can be reused and adapted, reducing engineering overhead and accelerating the prototyping cycle. Thus, we introduce the following four reward terms that are generalizable enough to solve the aforementioned three tasks with the same set of scaling factors.

\paragraph{Initial manipulation hand pose}
The initial hand pose relative to the object strongly impacts functional success~\cite{urain2023se, Jiang-RSS-21}. 
We define a task-specific initial hand pose for each hand and attach the pose onto the object, as shown in Fig.~\ref{fig:reward} (top-left). We thus encourage the pose alignment:
\begin{equation}
    r_{\text{align}} = - \alpha\big\| p_{\text{palm}} - p_{\text{def}} \big\|^2 - \beta d_{\text{quat}}(q_{\text{palm}}, q_{\text{def}}),
\end{equation}
where $p_{\text{palm}}$ and $q_{\text{palm}}$ are the current robot hand pose and $p_{\text{def}}$ and $q_{\text{def}}$ are the predefined initial hand pose depending on the object's initialization. $d_{\text{quat}}$ is the distance between two quaternions, and $\alpha$ and $\beta$ are scaling factors.
\paragraph{Object contact}
After aligning the robot hand with the initial manipulation pose, we encourage the hand to establish contact with the object. Specifically, we minimize the distance between the fingertips and the object’s center of mass. This term ensures reliable contact initiation, providing a stable starting point for subsequent manipulation:
\begin{equation}
    r_{\text{contact}} = \sum_{i=1}^n \frac{1}{1+\alpha^i\big\| p_{\text{tip}}^i - p_{\text{com}} \big\|^2},
\end{equation}
where $n$ is the number of fingertips, $p_{\text{tip}}^i$ is the position of the $i$-th fingertip, $p_{\text{com}}$ denotes the object center of mass, and $\alpha^i$ is the corresponding fingertip weight. 
\paragraph{Object goal poses with lifting gate}
Objects typically have designated target poses associated with task completion, e.g., placing an object into a box. We therefore define a goal pose for each object, where orientation constraints can be optionally included depending on task requirements. Prior works enforce full object pose trajectories, which are prone to errors when retargeting from human videos and often suboptimal under extensive pose randomization~\cite{yuan2025hermes, dan2025x, chen2025vividex, lum2025crossing}. Instead, we reward progress toward flexible goal poses, allowing the RL agent to explore diverse strategies for achieving the target. 
To prevent undesirable behaviors such as pushing objects rather than manipulating them in the air, we further introduce a lifting condition: the reward is only activated once the object is lifted above a height threshold:
\begin{equation}
r_{\text{goal}} =
\begin{cases}
    \max(d_z, 0), & \text{if } d_z < \tau, \\[6pt]
    \exp{(- \alpha \| p_{\text{o}} - p_{\text{g}} \|^2 
    - \beta \, d_{\text{quat}}(q_{\text{o}}, q_{\text{g}}))}, & \text{otherwise.}
\end{cases}
\end{equation}
where $d_z = |p_{\text{o},z} - \hat{p}_{\text{o},z}|$ and $\hat{p}_{\text{o},z}$ is the height of initial object position and $\tau$ is the lifting threshold.

\paragraph{Robot reset}
We further include a reset term that encourages the robot arm to return to a predefined configuration once the sub-task is completed, e.g., the object is already inside the drawer (Fig.~\ref{fig:reward} (bottom-right)). This term not only prevents the robot from executing irrelevant or unsafe behaviors during bimanual interaction but also provides a consistent starting state that facilitates chaining of tasks and smooth integration into long-horizon tasks:
\begin{equation}
    r_\text{reset} = \exp{(- \alpha \| \theta - \theta_{\text{reset}} \|^2)},
\end{equation}
where $\theta_{\text{reset}}$ is the reset joint configuration. 

\begin{figure}[t]
    \centering \includegraphics[width=.95\linewidth]{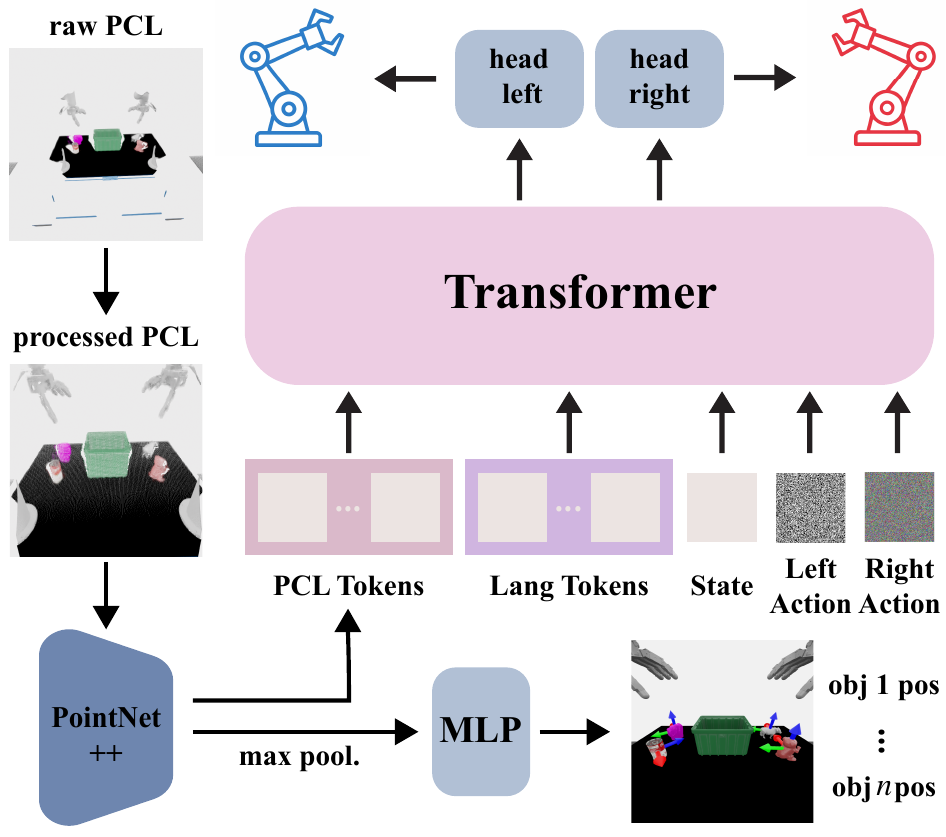}
    \caption{Network architecture. The raw point cloud is cropped to the region of interest and downsampled. Point cloud features, language tokens, robot states, and noisy actions are processed by a transformer, after which separate heads predict the action noises for the right and left hands. An additional MLP predicts the ground-truth object states.}
    \label{fig:network}
\end{figure}
\begin{figure*}[t]
    \centering
    \includegraphics[width=\linewidth]{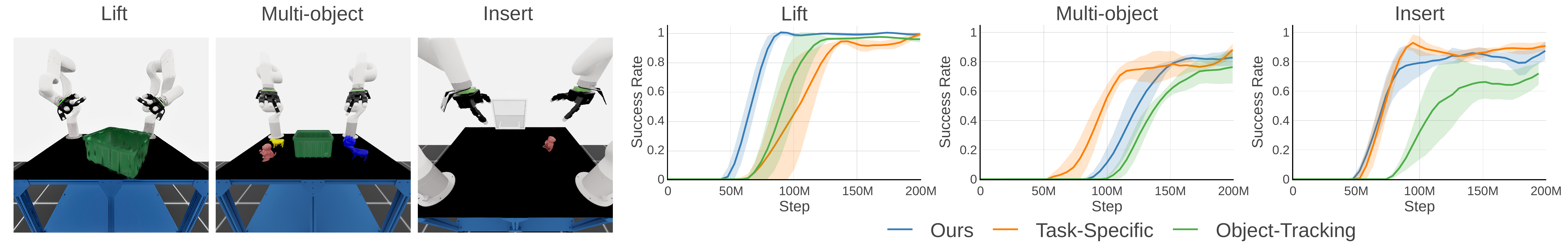}
    \caption{(Left) Visualization of three bimanual dexterous manipulation tasks in simulation. (Right) Performance of the proposed reward design compared to baselines. Our rewards achieve performance comparable to task-specific designs and outperform the object reference trajectory baseline.}
    \label{fig:reward_performance}
\end{figure*}

\subsection{Policy Training and Multimodal Data Collection}
\label{sec:method:b}
We obtain expert policies for each simulation task by optimizing the proposed generalizable rewards. Learning even a single bimanual dexterous manipulation task is challenging due to the high-dimensional observation and action spaces and the inherent difficulty of credit assignment. To address this, we formulate the bimanual problem as a multi-agent setting and adopt independent PPO (IPPO)~\cite{schulman2017proximal}, following \cite{li2025morphologically}. Each arm is controlled by a dedicated policy conditioned on its local observations, with its own value function defined over the local observations and rewards. This decomposition reduces exploration space, simplifies credit assignment, and improves learning efficiency. During teacher policy training, we apply domain randomization to factors that directly affect state-based learning, such as object initialization and physical parameters. Visual attributes are randomized only during subsequent data collection described below.

Once the RL teacher policies converge, we construct a diverse multimodal dataset by rolling them out in simulation. To maximize diversity, we employ extensive domain randomization as shown in Fig.~\ref{fig:overview}(2). For objects, we draw from the dataset in \cite{chen2023visual}, which includes a wide range of complex and nonconvex geometries such as cars, shoes, and toys. Each object is randomized in pose, physical parameters (e.g., static and dynamic friction, mass, restitution), and visual attributes (e.g., color, reflectivity, roughness). Specifically, we select $100$ different objects and process them into similar scale that are suitable for grasping, and $20$ color categories. We also randomize perception-related factors including camera viewpoints, intrinsic parameters, and lighting conditions. This combination of physical and visual perturbations exposes the policy to a broad distribution of scenarios, enhancing its robustness, and improving the transfer to real-world settings.

\subsection{Automated Language Labeling}
\label{sec:method:c}
One major advantage of simulation-based data generation is the ability to automatically generate language labels for every trajectory~\cite{chen2025robotwin}. We predefine language annotations for objects, colors, and goals, and then instantiate task-specific descriptions at each environment reset using a template format. For instance, in the multi-object pick-and-place task in Fig.~\ref{fig:overview}(3), we generate the instruction: “use the right arm to place the red car into the bin, and the left arm to place the blue shoe into the bin.” To increase linguistic diversity, we employ GPT~\cite{hurst2024gpt} to paraphrase the templates: “pick the nearest objects with both arms and place them respectively.” The augmentation enriches the dataset with natural language variability, improving the generalization ability to unseen instructions. The language input is primarily used to disambiguate task goals, object choices, and arm-object assignments within the shared multi-task policy, rather than to improve single-task imitation in isolation.

\subsection{Multi-task Policy Distillation}
\label{sec:method:d}
After constructing a large-scale multimodal dataset, our goal is to train a multi-task policy capable of executing diverse tasks conditioned on language instructions. Exploiting such large-scale synthetic data is challenging, so we model the policy as a transformer-based diffusion model due to its expressive capacity. The denoising network $\epsilon_\theta$ takes as input a 6-channel point cloud $O$ along with the robots' state $S$ and language tokens $L$ and predicts the noise $\epsilon^k$ required for the diffusion process, where $\epsilon^k$ is the noise at $k$ iteration by doing forward diffusion process at a data point $a^0$. The action loss is defined as $\ell_{\text{action}} = \big\| \epsilon^k - \epsilon_\theta(\bar{\alpha}_ka^0+\bar{\beta}_k\epsilon^k, k, O, S, L) \big\|^2$, where $\bar{\alpha}_k$ and $\bar{\beta}_k$ are noise schedulers. In practice, we train with $20$ denoising steps and adopt $5$ inference steps using the DDIM~\cite{song2020denoising} sampler. 

\paragraph{Network architecture (Fig.\ref{fig:network}).}


We first preprocess the raw point cloud by cropping it to the workspace region of interest (e.g., the tabletop) and randomly downsampling it to $2,048$ points. 
The processed points are encoded using a PointNet++~\cite{qi2017pointnet++} encoder without max pooling to preserve spatial structure, and the resulting point features are projected onto a 256-dimensional embedding space. 
In parallel, the low-level robot state, including joint positions and velocities, is embedded using Fourier positional encodings to improve temporal discriminability across steps.  

Bimanual manipulation often requires the two arms to perform actions that are partially independent yet coordinated~\cite{jiang2025rethinking}. 
To reflect this structure, we decouple the action representation into separate queries for each side. 
Through cross-attention, these action queries interact with the conditioning tokens, which consist of point cloud features, robot state features, and language embeddings encoded via SigLip~\cite{zhai2023sigmoid}, at linear complexity $O(n)$ rather than quadratic $O(n^2)$.  
Finally, two output heads decode the refined action tokens to predict the diffusion noise for each arm. 

\paragraph{Auxiliary loss for state prediction.}
A unique advantage of simulation-based data is the availability of ground-truth object states, which can serve as anchors for perception. We  leverage this opportunity by introducing an auxiliary prediction head that maps the point cloud features to object states $\ell_{\text{aux}} = \big\| \hat{x}_{\text{obj}} -x_{\text{obj}} \big\|^2$, where $x_{\text{obj}}$ is the object state including positions and orientations and $\hat{x}_{\text{obj}}$ is the prediction. Note that for the continuity of the rotation and smoothness of learning, we adopt 6D representations~\cite{zhou2019continuity}. This auxiliary supervision regularizes the visual encoder, encouraging it to extract task-relevant geometric information and improving policy robustness in vision-conditioned settings.

\section{EXPERIMENTS}
\label{sec:exp}
\subsection{Real-World and Simulation Setup}
\label{sec:exp:a}
Our robotic platform consists of two 6-degree-of-freedom (DOF) xArm UF850 arms, each equipped with a 16-DOF Allegro Hand V4, yielding a total of 44 degrees of freedom. The system operates over a 1.2$\times$0.8m tabletop workspace under standard safety constraints. A low-level joint-level PD controller runs at 120Hz, while policy inference is executed at 20Hz. Perception is provided by a single egocentric ZED2i RGB-D camera mounted in-between the arms. We use the NVIDIA IsaacLab~\cite{mittal2023orbit} as our simulator and construct a digital twin that accurately replicates the real-world setup. We follow \cite{lin2025sim} to identify the physics parameters in simulation to ensure smooth sim-to-real transfer.

\begin{figure*}[t]
    \centering
    \includegraphics[width=\linewidth]{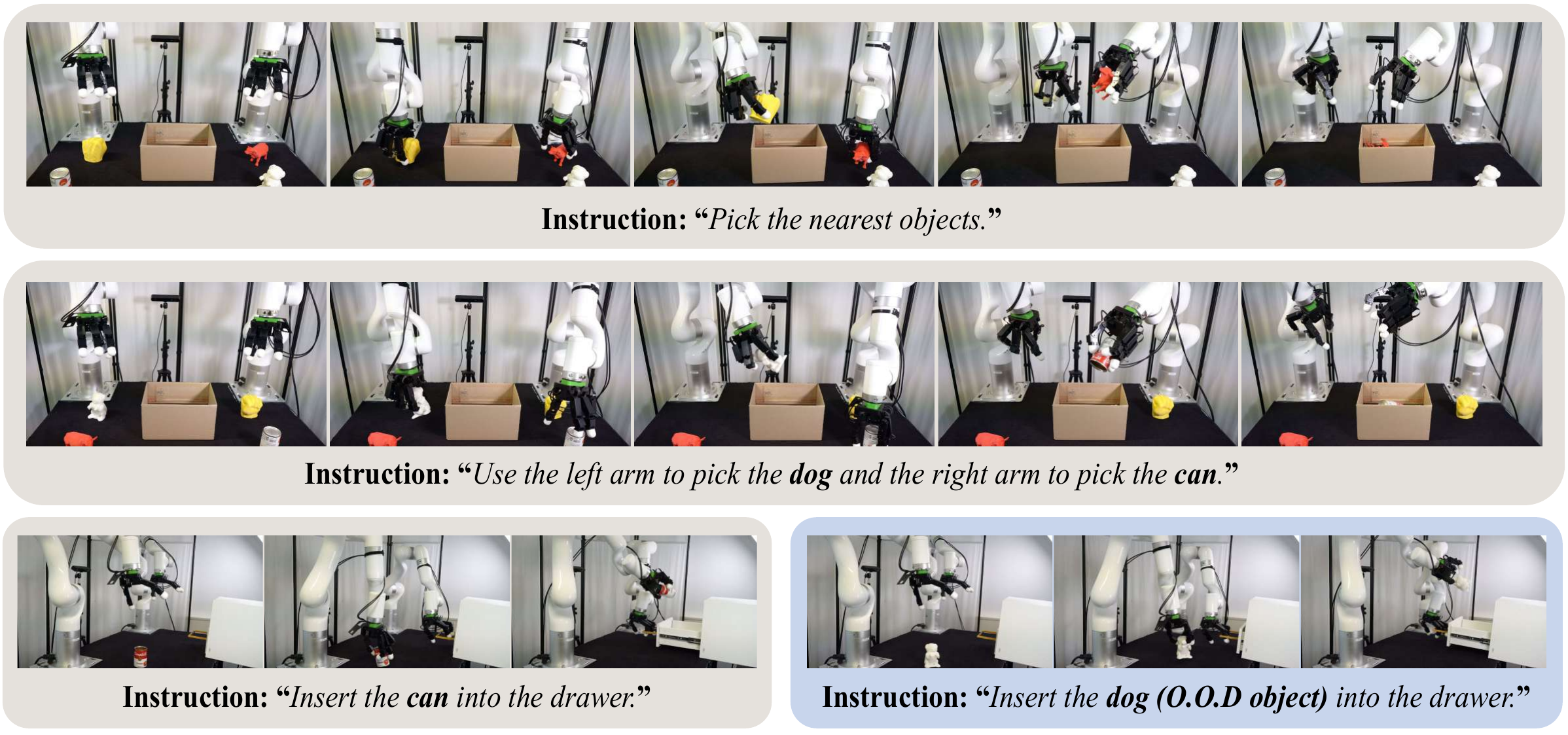}
    \caption{Visualization of real-world results. The first two rows show the \emph{Multi-Object} task under different language instructions, while the last row illustrates the \emph{Insert} task with both in-domain and out-of-domain objects.}
    \label{fig:real}
\end{figure*}

\subsection{Evaluation of Generalizable Reward Design}
\label{sec:exp:b}
To construct datasets across diverse tasks, the first question to answer is whether the proposed generalizable reward design can achieve high success rates and transfer effectively across different manipulation tasks. We compare our reward design against two baselines: (i) task-specific rewards manually designed for each task as in \cite{li2025morphologically}, and (ii) the object reference trajectory reward from \cite{chen2025vividex}, a widely adopted approach in recent work (see Section~\ref{sec:related:b}). Note that we use the \textbf{same reward terms and scaling factors} for all tasks, emphasizing the scalability for reward tuning. 

\textbf{Learning efficiency and success rate.}
Our proposed reward design achieves comparable learning efficiency and overall success rates compared to the task-specific reward. As shown in Fig.~\ref{fig:reward_performance}, it performs comparably on the \emph{Insert} task and outperforms it on \emph{Lift}. We attribute this improvement to the fact that overly constrained task-specific rewards may hinder the learning, whereas simpler rewards allow more flexible exploration. On the \emph{Multi-Object} task, however, our performance is lower than with task-specific rewards because aligning hand poses with multiple objects introduces constraints that slow down learning. This suggests that our four reward terms can be selectively combined depending on the task to relax the constraints. By contrast, the object reference trajectory reward performs poorly across all tasks. The key limitation is its reliance on predefined trajectories: when objects are heavily randomized, the robot often encounters joint limits that while following the trajectory. These results strengthen our key idea: instead of forcing adherence to a fixed trajectory, it is more effective to provide flexible goal conditions that allow the RL agent to discover.

\textbf{Implementation efficiency.}
In addition to performance, scalability requires efficient reward design across tasks. Empirically, the most time-consuming component of our approach is defining the initial hand alignment pose. Since only one pose per hand is required, reward setup takes roughly one hour per task. In contrast, setting up object reference trajectory rewards can take up to a full day, as it involves video recording, object tracking, and coordinate alignment between the video and the simulator. 

\subsection{Evaluation of Policy Distillation}
We evaluate the distilled vision-based policy in both simulation and real-world settings, considering both single-task and multi-task variants. Each task has equal percentage in the constructed multi-task dataset. Simulation results are obtained with $256$ parallel environments, while real-world performance is measured over $20$ trials per experiment.

\begin{table}[]
\centering
\begin{tabular}{l|c|c|c|c|c}
\toprule
   & \textit{Multi-Object} & \textit{Insert} & \textit{Lift}  & \textit{Multi-Task} \\
\midrule
DP3~\cite{ze20243d}  &              $0.42$                   &       $0.38$            &     $0.703$                     &      $0.06$              \\
iDP3~\cite{ze2024generalizable} &         $0.41$             &        $0.47$             &      $0.65$                     &    $0.14$                 \\
Ours &       0.80                  &        0.74               &      $\textbf{0.96}$           &       0.67             \\     
w/o sep. head & \textbf{0.812}  & \textbf{0.79} &  0.93  &  \textbf{0.70} \\
w/o state pre.   &         $0.72$              &     $0.68$             &        $0.85$          &         $0.57$               \\
\bottomrule
\end{tabular}
\vskip 1mm
\begin{tabular}{l|c|c|c|c|c}
\toprule
 & \textit{Multi-Object} & \textit{Insert} & \textit{Lift}  & \textit{Multi-Task} \\
\midrule
DP3~\cite{ze20243d}   &              $10/20$                 &       $5/20$          &       $10/20$           &       $4/20$            \\
iDP3~\cite{ze2024generalizable}          &        $8/20$             &             $3/20$        &        $11/20$     &           $2/20$          \\
Ours                   &       $\textbf{15/20}$              &              $\textbf{16/20}$          &        $14/20$           &         $\textbf{8/20}$           \\
w/o sep. head    &             $11/20$              &             $8/20$          &        $\textbf{16/20}$        &           $2/20$         \\
w/o state pre.          &      $10/20$                     &        $12/20$       &        $8/20$         &            $3/20$           \\
\bottomrule
\end{tabular}
\caption{(Top) Simulation and (Bottom) real-world results comparing baselines DP3~\cite{ze20243d} and iDP3~\cite{ze2024generalizable} and ablated design choices.}
\label{tab:sim}
\end{table}

\textbf{Comparison with SOTA baselines.}
We compare our approach against two recent point-cloud-based policies, DP3~\cite{ze20243d} and iDP3~\cite{ze2024generalizable}. The key architectural differences lie in the point cloud encoder: DP3 employs a linear encoder, iDP3 uses a pyramid convolutional encoder, and our method adopts a PointNet++ encoder (without max pooling) to better preserve spatial features. For fairness, we standardize the input size to $2,048$ points, which lies between the settings used in the baselines ($1,024$ and $4,096$). As shown in Tab.~\ref{tab:sim}, our policy consistently outperforms both DP3 and iDP3 in single-task and multi-task settings. Interestingly, the performance gap between DP3 and iDP3 is smaller than reported in their original papers. 
We attribute this to differences in data: our dataset, with on the order of $10^3$ trajectories collected from converged RL policies, is orders of magnitude larger and more consistent than human-teleoperated datasets with only $10^1$ trajectories, which likely reduces sensitivity to architectural differences.

In Fig.~\ref{fig:real}, we visualize real-world rollouts, showing that the multi-task policy successfully follows diverse language instructions, enabled by the augmented language labels. Furthermore, in the \emph{Insert} task, we evaluate an out-of-distribution object (a dog) that was excluded from the \emph{Insert} training set but included in the \emph{Multi-Object} task. Due to the multi-task training paradigm, the policy effectively transfers knowledge from the multi-object dataset to the novel task, highlighting the potential of learning multi-task policies from specialized RL teachers.

\textbf{Ablations on design choices.}
We further investigate two design choices: (i) the use of separate action heads and (ii) the state-prediction auxiliary loss. As shown in Tab.~\ref{tab:sim}, removing separate action heads and using a single shared head achieves higher success rates, we hypothesize that it is easier to fit the training distribution. However, in the real world, the separated heads yield significantly higher success rates except \emph{Lift}. In practice, we observe that independent heads increase robustness to out-of-distribution cases arising from sim-to-real gaps. For example, if one hand grasps an object with an unexpected behavior, the other hand remains unaffected, allowing the policy to recover more effectively. However, since \emph{Lift} is a single-object manipulation task, both hands will be affected if the box is tilted and therefore a shared observation and action space is preferred. Incorporating the state-prediction auxiliary loss consistently improves performance across tasks. This leverages simulation’s unique advantage of providing ground-truth object states, effectively anchoring the visual encoder to task-relevant features. We view this as a ``cherry on the cake”, and recommend including this auxiliary signal whenever policies are distilled from simulation data.

\begin{table}[h]
\centering
\begin{tabular}{p{1.6cm}|cc|cc|cc}
\toprule
\multirow{2}{*}{Dataset Size} & \multicolumn{2}{c|}{\textit{Multi-Object}} & \multicolumn{2}{c|}{\textit{Insert}} & \multicolumn{2}{c}{\textit{Multi-Task}} \\
                  & Sim                 & Real                & Sim              & Real             & Sim                & Real               \\
\midrule
$500$              &      $0.590$                &    $9/20$                 &      $0.351$            &       $5/20$           &          $0.398$          &        $3/20$            \\
$2,000$    &       $0.758$               &        $13/20$             &         $0.539$         &       $7/20$           &    $0.519$                &       $3/20$             \\
$5,000$  &        $\textbf{0.801}$             &         $\textbf{15/20}$            &        $\textbf{0.738}$          &       $\textbf{16/20}$            &             $\textbf{0.672}$       &        $\textbf{8/20}$           \\
\bottomrule
\end{tabular}
\caption{Results of policies trained with varying dataset sizes.}
\label{tab:real}
\end{table}

\textbf{Impact of dataset size.}
As shown in Tab.~\ref{tab:real}, our transformer-based architecture scales effectively as the dataset size increases. We attribute this to both the model capacity of the transformer and the underlying data distribution within the dataset. Since only successful trajectories are collected, corner-case initializations are more likely to fail, leading to an imbalanced dataset. With larger datasets, these rare cases are better included, allowing the student policy to imitate more robust strategies. We also observe that the effect of dataset size depends on task complexity. For example, the \emph{Multi-Object} task is relatively straightforward, requiring less inter-arm coordination and no interaction with articulated objects. In contrast, the \emph{Insert} task is more complex, and therefore suffers greater performance degradation as dataset size decreases. This finding highlights both the importance of dataset scale and the critical role of large-scale synthetic data generation. 

\section{CONCLUSIONS}

In this work, we proposed RLDC, a scalable pipeline that leverages reinforcement learning to generate high-quality, diverse training data for bimanual dexterous manipulation. The scale and diversity of training data are increasingly critical for developing multi-task and foundation models in robotics. In response, we show that synthetic data generated by RL with extensive randomization and automatic language labeling can efficiently produce large-scale multimodal datasets. Unlike direct augmentation of human demonstrations, RL-generated data are collected from closed-loop robot policies in simulation, which encourages consistency with robot kinematics and task feasibility under randomized conditions.

To enable scalable RL-based data collection, we introduced several innovations: (1) a generalizable reward function tailored to bimanual dexterous manipulation reduces engineering overhead and accelerates policy prototyping; (2) converged teacher policies are rolled out to collect diverse trajectories with extensive randomization and grounded language instructions; and (3) the dataset is distilled into a diffusion-based student policy with visual anchoring for robust sim-to-real transfer. Experiments show that our reward design achieves success rates comparable to finely tuned task-specific rewards and that multi-task policies can be learned entirely from synthetic data with strong real-world transfer.

\textbf{Limitations and future work.}
Our study has two key limitations. First, setting up simulation environments and initial hand poses remain time-consuming. While reward engineering accounts for much of this cost, additional effort is required to tune robot physics parameters, prepare assets, and define success criteria. Recent work~\cite{lin2025sim} suggests that parallel simulation can accelerate physics parameter tuning. Community resources such as \cite{chen2025robotwin, nasiriany2024robocasa, geng2025roboverse, srivastava2022behavior} increasingly provide reusable assets and environments to mitigate this burden, and another line of work~\cite{dan2025x, torne2024reconciling} focuses on automating real-to-sim that rapidly reconstruct real-world environments in simulation. Second, RLDC inherits the behavior biases of its RL teachers. Although the resulting trajectories are robot-feasible, they are not necessarily human-like and can contain unnecessary finger motions or abrupt releases. This limitation is especially relevant for delicate objects and tasks requiring controlled placement, fine in-hand regrasping, finger gaiting, or deformable-object interaction. Future work could combine RL-generated trajectories with human demonstrations, smoothness/contact regularization, or release-quality constraints to obtain both feasible and natural behaviors.









\bibliographystyle{IEEEtran}
\bibliography{main}

@article{chen2025robotwin,
  title={RoboTwin 2.0: A Scalable Data Generator and Benchmark with Strong Domain Randomization for Robust Bimanual Robotic Manipulation},
  author={Chen, Tianxing and Chen, Zanxin and Chen, Baijun and Cai, Zijian and Liu, Yibin and Liang, Qiwei and Li, Zixuan and Lin, Xianliang and Ge, Yiheng and Gu, Zhenyu and others},
  journal={arXiv preprint arXiv:2506.18088},
  year={2025}
}

@inproceedings{zhai2023sigmoid,
  title={Sigmoid loss for language image pre-training},
  author={Zhai, Xiaohua and Mustafa, Basil and Kolesnikov, Alexander and Beyer, Lucas},
  booktitle={Proceedings of the IEEE/CVF international conference on computer vision},
  pages={11975--11986},
  year={2023}
}

@article{qi2017pointnet++,
  title={Pointnet++: Deep hierarchical feature learning on point sets in a metric space},
  author={Qi, Charles Ruizhongtai and Yi, Li and Su, Hao and Guibas, Leonidas J},
  journal={Advances in neural information processing systems},
  volume={30},
  year={2017}
}

@inproceedings{wang2024cyberdemo,
  title={Cyberdemo: Augmenting simulated human demonstration for real-world dexterous manipulation},
  author={Wang, Jun and Qin, Yuzhe and Kuang, Kaiming and Korkmaz, Yigit and Gurumoorthy, Akhilan and Su, Hao and Wang, Xiaolong},
  booktitle={Proceedings of the IEEE/CVF Conference on Computer Vision and Pattern Recognition},
  pages={17952--17963},
  year={2024}
}

@inproceedings{jiang2025dexmimicgen,
  title={Dexmimicgen: Automated data generation for bimanual dexterous manipulation via imitation learning},
  author={Jiang, Zhenyu and Xie, Yuqi and Lin, Kevin and Xu, Zhenjia and Wan, Weikang and Mandlekar, Ajay and Fan, Linxi Jim and Zhu, Yuke},
  booktitle={2025 IEEE International Conference on Robotics and Automation (ICRA)},
  pages={16923--16930},
  year={2025},
  organization={IEEE}
}

@article{xu2024rldg,
  title={Rldg: Robotic generalist policy distillation via reinforcement learning},
  author={Xu, Charles and Li, Qiyang and Luo, Jianlan and Levine, Sergey},
  journal={arXiv preprint arXiv:2412.09858},
  year={2024}
}

@article{yuan2025hermes,
  title={HERMES: Human-to-Robot Embodied Learning from Multi-Source Motion Data for Mobile Dexterous Manipulation},
  author={Yuan, Zhecheng and Wei, Tianming and Gu, Langzhe and Hua, Pu and Liang, Tianhai and Chen, Yuanpei and Xu, Huazhe},
  journal={arXiv preprint arXiv:2508.20085},
  year={2025}
}

@article{eschmann2021reward,
  title={Reward function design in reinforcement learning},
  author={Eschmann, Jonas},
  journal={Reinforcement learning algorithms: Analysis and Applications},
  pages={25--33},
  year={2021},
  publisher={Springer}
}

@article{lin2025sim,
  title={Sim-to-real reinforcement learning for vision-based dexterous manipulation on humanoids},
  author={Lin, Toru and Sachdev, Kartik and Fan, Linxi and Malik, Jitendra and Zhu, Yuke},
  journal={arXiv preprint arXiv:2502.20396},
  year={2025}
}

@article{dan2025x,
  title={X-Sim: Cross-Embodiment Learning via Real-to-Sim-to-Real},
  author={Dan, Prithwish and Kedia, Kushal and Chao, Angela and Duan, Edward Weiyi and Pace, Maximus Adrian and Ma, Wei-Chiu and Choudhury, Sanjiban},
  journal={arXiv preprint arXiv:2505.07096},
  year={2025}
}

@article{lum2025crossing,
  title={Crossing the human-robot embodiment gap with sim-to-real rl using one human demonstration},
  author={Lum, Tyler Ga Wei and Lee, Olivia Y and Liu, C Karen and Bohg, Jeannette},
  journal={arXiv preprint arXiv:2504.12609},
  year={2025}
}

@inproceedings{chen2025vividex,
  title={Vividex: Learning vision-based dexterous manipulation from human videos},
  author={Chen, Zerui and Chen, Shizhe and Arlaud, Etienne and Laptev, Ivan and Schmid, Cordelia},
  booktitle={2025 IEEE International Conference on Robotics and Automation (ICRA)},
  pages={3336--3343},
  year={2025},
  organization={IEEE}
}

@article{singh2024dextrah,
  title={Dextrah-rgb: Visuomotor policies to grasp anything with dexterous hands},
  author={Singh, Ritvik and Allshire, Arthur and Handa, Ankur and Ratliff, Nathan and Van Wyk, Karl},
  journal={arXiv preprint arXiv:2412.01791},
  year={2024}
}

@article{lum2024dextrah,
  title={Dextrah-g: Pixels-to-action dexterous arm-hand grasping with geometric fabrics},
  author={Lum, Tyler Ga Wei and Matak, Martin and Makoviychuk, Viktor and Handa, Ankur and Allshire, Arthur and Hermans, Tucker and Ratliff, Nathan D and Van Wyk, Karl},
  journal={arXiv preprint arXiv:2407.02274},
  year={2024}
}

@article{yuan2024learning,
  title={Learning to manipulate anywhere: A visual generalizable framework for reinforcement learning},
  author={Yuan, Zhecheng and Wei, Tianming and Cheng, Shuiqi and Zhang, Gu and Chen, Yuanpei and Xu, Huazhe},
  journal={arXiv preprint arXiv:2407.15815},
  year={2024}
}

@inproceedings{chen2023visual,
  title     = {Visual dexterity: In-hand dexterous manipulation from depth},
  author    = {Chen, Tao and Tippur, Megha and Wu, Siyang and Kumar, Vikash and Adelson, Edward and Agrawal, Pulkit},
  booktitle = {Icml workshop on new frontiers in learning, control, and dynamical systems},
  year      = {2023}
}

@inproceedings{lin2024twisting,
  title     = {Twisting Lids Off with Two Hands},
  author    = {Toru Lin and Zhao-Heng Yin and Haozhi Qi and Pieter Abbeel and Jitendra Malik},
  booktitle = {8th Annual Conference on Robot Learning (CoRL)},
  year      = {2024},
  url       = {https://openreview.net/forum?id=3wBqoPfoeJ}
}

@article{li2025morphologically,
  title={Morphologically Symmetric Reinforcement Learning for Ambidextrous Bimanual Manipulation},
  author={Li, Zechu and Jin, Yufeng and Apraez, Daniel Ordonez and Semini, Claudio and Liu, Puze and Chalvatzaki, Georgia},
  journal={arXiv preprint arXiv:2505.05287},
  year={2025}
}

@article{jiang2025rethinking,
  title={Rethinking Bimanual Robotic Manipulation: Learning with Decoupled Interaction Framework},
  author={Jiang, Jian-Jian and Wu, Xiao-Ming and He, Yi-Xiang and Zeng, Ling-An and Wei, Yi-Lin and Zhang, Dandan and Zheng, Wei-Shi},
  journal={arXiv preprint arXiv:2503.09186},
  year={2025}
}

@inproceedings{zhou2019continuity,
  title={On the continuity of rotation representations in neural networks},
  author={Zhou, Yi and Barnes, Connelly and Lu, Jingwan and Yang, Jimei and Li, Hao},
  booktitle={Proceedings of the IEEE/CVF conference on computer vision and pattern recognition},
  pages={5745--5753},
  year={2019}
}

@article{schulman2017proximal,
  title   = {Proximal policy optimization algorithms},
  author  = {Schulman, John and Wolski, Filip and Dhariwal, Prafulla and Radford, Alec and Klimov, Oleg},
  journal = {arXiv preprint arXiv:1707.06347},
  year    = {2017}
}

@article{mittal2023orbit,
  author  = {Mittal, Mayank and Yu, Calvin and Yu, Qinxi and Liu, Jingzhou and Rudin, Nikita and Hoeller, David and Yuan, Jia Lin and Singh, Ritvik and Guo, Yunrong and Mazhar, Hammad and Mandlekar, Ajay and Babich, Buck and State, Gavriel and Hutter, Marco and Garg, Animesh},
  journal = {IEEE Robotics and Automation Letters},
  title   = {Orbit: A Unified Simulation Framework for Interactive Robot Learning Environments},
  year    = {2023},
  volume  = {8},
  number  = {6},
  pages   = {3740-3747},
  doi     = {10.1109/LRA.2023.3270034}
}

@article{black2024pi_0,
  title={$\pi_0$: A Vision-Language-Action Flow Model for General Robot Control},
  author={Black, Kevin and Brown, Noah and Driess, Danny and Esmail, Adnan and Equi, Michael and Finn, Chelsea and Fusai, Niccolo and Groom, Lachy and Hausman, Karol and Ichter, Brian and others},
  journal={arXiv preprint arXiv:2410.24164},
  year={2024}
}

@article{kim2024openvla,
  title={Openvla: An open-source vision-language-action model},
  author={Kim, Moo Jin and Pertsch, Karl and Karamcheti, Siddharth and Xiao, Ted and Balakrishna, Ashwin and Nair, Suraj and Rafailov, Rafael and Foster, Ethan and Lam, Grace and Sanketi, Pannag and others},
  journal={arXiv preprint arXiv:2406.09246},
  year={2024}
}

@article{team2024octo,
  title={Octo: An open-source generalist robot policy},
  author={Team, Octo Model and Ghosh, Dibya and Walke, Homer and Pertsch, Karl and Black, Kevin and Mees, Oier and Dasari, Sudeep and Hejna, Joey and Kreiman, Tobias and Xu, Charles and others},
  journal={arXiv preprint arXiv:2405.12213},
  year={2024}
}

@article{wen2025tinyvla,
  title={Tinyvla: Towards fast, data-efficient vision-language-action models for robotic manipulation},
  author={Wen, Junjie and Zhu, Yichen and Li, Jinming and Zhu, Minjie and Tang, Zhibin and Wu, Kun and Xu, Zhiyuan and Liu, Ning and Cheng, Ran and Shen, Chaomin and others},
  journal={IEEE Robotics and Automation Letters},
  year={2025},
  publisher={IEEE}
}

@article{reuss2025flower,
  title={Flower: Democratizing generalist robot policies with efficient vision-language-action flow policies},
  author={Reuss, Moritz and Zhou, Hongyi and R{\"u}hle, Marcel and Ya{\u{g}}murlu, {\"O}mer Erdin{\c{c}} and Otto, Fabian and Lioutikov, Rudolf},
  journal={arXiv preprint arXiv:2509.04996},
  year={2025}
}

@article{chi2024diffusionpolicy,
	author = {Cheng Chi and Zhenjia Xu and Siyuan Feng and Eric Cousineau and Yilun Du and Benjamin Burchfiel and Russ Tedrake and Shuran Song},
	title ={Diffusion Policy: Visuomotor Policy Learning via Action Diffusion},
	journal = {The International Journal of Robotics Research},
	year = {2024},
}

@inproceedings{fu2024mobile,
  author    = {Fu, Zipeng and Zhao, Tony Z. and Finn, Chelsea},
  title     = {Mobile ALOHA: Learning Bimanual Mobile Manipulation with Low-Cost Whole-Body Teleoperation},
  booktitle = {{Conference on Robot Learning (CoRL)}},
  year      = {2024},
}

@article{funk2024actionflow,
  title={Actionflow: Equivariant, accurate, and efficient policies with spatially symmetric flow matching},
  author={Funk, Niklas and Urain, Julen and Carvalho, Joao and Prasad, Vignesh and Chalvatzaki, Georgia and Peters, Jan},
  journal={arXiv preprint arXiv:2409.04576},
  year={2024}
}

@inproceedings{urain2023se,
  title={SE (3)-DiffusionFields: Learning smooth cost functions for joint grasp and motion optimization through diffusion},
  author={Urain, Julen and Funk, Niklas and Peters, Jan and Chalvatzaki, Georgia},
  booktitle={2023 IEEE International Conference on Robotics and Automation (ICRA)},
  pages={5923--5930},
  year={2023},
  organization={IEEE}
}

@INPROCEEDINGS{Jiang-RSS-21, 
    AUTHOR    = {Zhenyu Jiang AND Yifeng Zhu AND Maxwell Svetlik AND Kuan Fang AND Yuke Zhu}, 
    TITLE     = {{Synergies Between Affordance and Geometry: 6-DoF Grasp Detection via Implicit Representations}}, 
    BOOKTITLE = {Proceedings of Robotics: Science and Systems}, 
    YEAR      = {2021}, 
    ADDRESS   = {Virtual}, 
    MONTH     = {July}, 
    DOI       = {10.15607/RSS.2021.XVII.024} 
}

@article{ze20243d,
  title={3d diffusion policy: Generalizable visuomotor policy learning via simple 3d representations},
  author={Ze, Yanjie and Zhang, Gu and Zhang, Kangning and Hu, Chenyuan and Wang, Muhan and Xu, Huazhe},
  journal={arXiv preprint arXiv:2403.03954},
  year={2024}
}

@article{ze2024generalizable,
  title={Generalizable humanoid manipulation with 3d diffusion policies},
  author={Ze, Yanjie and Chen, Zixuan and Wang, Wenhao and Chen, Tianyi and He, Xialin and Yuan, Ying and Peng, Xue Bin and Wu, Jiajun},
  journal={arXiv preprint arXiv:2410.10803},
  year={2024}
}

@article{nasiriany2024robocasa,
  title={Robocasa: Large-scale simulation of everyday tasks for generalist robots},
  author={Nasiriany, Soroush and Maddukuri, Abhiram and Zhang, Lance and Parikh, Adeet and Lo, Aaron and Joshi, Abhishek and Mandlekar, Ajay and Zhu, Yuke},
  journal={arXiv preprint arXiv:2406.02523},
  year={2024}
}

@article{geng2025roboverse,
  title={RoboVerse: Towards a unified platform, dataset and benchmark for scalable and generalizable robot learning},
  author={Geng, Haoran and Wang, Feishi and Wei, Songlin and Li, Yuyang and Wang, Bangjun and An, Boshi and Cheng, Charlie Tianyue and Lou, Haozhe and Li, Peihao and Wang, Yen-Jen and others},
  journal={arXiv preprint arXiv:2504.18904},
  year={2025}
}

@inproceedings{srivastava2022behavior,
  title={Behavior: Benchmark for everyday household activities in virtual, interactive, and ecological environments},
  author={Srivastava, Sanjana and Li, Chengshu and Lingelbach, Michael and Mart{\'\i}n-Mart{\'\i}n, Roberto and Xia, Fei and Vainio, Kent Elliott and Lian, Zheng and Gokmen, Cem and Buch, Shyamal and Liu, Karen and others},
  booktitle={Conference on robot learning},
  pages={477--490},
  year={2022},
  organization={PMLR}
}

@article{torne2024reconciling,
  title={Reconciling reality through simulation: A real-to-sim-to-real approach for robust manipulation},
  author={Torne, Marcel and Simeonov, Anthony and Li, Zechu and Chan, April and Chen, Tao and Gupta, Abhishek and Agrawal, Pulkit},
  journal={arXiv preprint arXiv:2403.03949},
  year={2024}
}

@article{qin2023anyteleop,
  title={Anyteleop: A general vision-based dexterous robot arm-hand teleoperation system},
  author={Qin, Yuzhe and Yang, Wei and Huang, Binghao and Van Wyk, Karl and Su, Hao and Wang, Xiaolong and Chao, Yu-Wei and Fox, Dieter},
  journal={arXiv preprint arXiv:2307.04577},
  year={2023}
}

@article{qiu2025humanoid,
  title={Humanoid Policy Human Policy},
  author={Qiu, Ri-Zhao and Yang, Shiqi and Cheng, Xuxin and Chawla, Chaitanya and Li, Jialong and He, Tairan and Yan, Ge and Yoon, David J and Hoque, Ryan and Paulsen, Lars and others},
  journal={arXiv preprint arXiv:2503.13441},
  year={2025}
}

@article{song2020denoising,
  title={Denoising diffusion implicit models},
  author={Song, Jiaming and Meng, Chenlin and Ermon, Stefano},
  journal={arXiv preprint arXiv:2010.02502},
  year={2020}
}

@inproceedings{qi2023hand,
  title        = {In-hand object rotation via rapid motor adaptation},
  author       = {Qi, Haozhi and Kumar, Ashish and Calandra, Roberto and Ma, Yi and Malik, Jitendra},
  booktitle    = {Conference on Robot Learning},
  pages        = {1722--1732},
  year         = {2023},
  organization = {PMLR}
}

@article{chen2024vegetable,
  title   = {Vegetable peeling: A case study in constrained dexterous manipulation},
  author  = {Chen, Tao and Cousineau, Eric and Kuppuswamy, Naveen and Agrawal, Pulkit},
  journal = {arXiv preprint arXiv:2407.07884},
  year    = {2024}
}

@article{hurst2024gpt,
  title={Gpt-4o system card},
  author={Hurst, Aaron and Lerer, Adam and Goucher, Adam P and Perelman, Adam and Ramesh, Aditya and Clark, Aidan and Ostrow, AJ and Welihinda, Akila and Hayes, Alan and Radford, Alec and others},
  journal={arXiv preprint arXiv:2410.21276},
  year={2024}
}

\end{document}